\pdfoutput=1
\documentclass[12pt]{iopart}

\usepackage{hyperref}
\expandafter\let\csname equation*\endcsname\relax
\expandafter\let\csname endequation*\endcsname\relax
\usepackage{mathtools}
\usepackage{graphicx}
\usepackage{amssymb}
\usepackage{booktabs}
\usepackage{subfig}
\usepackage{subcaption} 

\begin{document}

\title[DRACO: Differentiable Reconstruction for Arbitrary CBCT Orbits]{DRACO: Differentiable Reconstruction for Arbitrary CBCT Orbits}

\author{Chengze~Ye,
        Linda-Sophie~Schneider,
        Yipeng~Sun,
        Mareike~Thies,
        Siyuan Mei,
        and~Andreas~Maier}

\address{Friedrich-Alexander University Erlangen-Nuremberg, Erlangen, Germany}
\ead{chengze.ye@fau.de}
\vspace{10pt}
\begin{indented}
\item[]October 2024
\end{indented}

\begin{abstract}
This paper introduces a novel method for reconstructing cone beam computed tomography (CBCT) images for arbitrary orbits using a differentiable shift-variant filtered backprojection (FBP) neural network. Traditional CBCT reconstruction methods for arbitrary orbits, like iterative reconstruction algorithms, are computationally expensive and memory-intensive. The proposed method addresses these challenges by employing a shift-variant FBP algorithm optimized for arbitrary trajectories through a deep learning approach that adapts to a specific orbit geometry.
This approach overcomes the limitations of existing techniques by integrating known operators into the learning model, minimizing the number of parameters, and improving the interpretability of the model. The proposed method is a significant advancement in interventional medical imaging, particularly for robotic C-arm CT systems, enabling faster and more accurate CBCT reconstructions with customized orbits. Especially this method can also be used for the analytical reconstruction of non-continuous orbits like circular plus arc.
The experimental results demonstrate that the proposed method significantly accelerates the reconstruction process compared to conventional iterative algorithms. It achieves comparable or superior image quality while reducing noise, as evidenced by metrics such as the mean squared error (MSE), the peak signal-to-noise ratio (PSNR), and the structural similarity index measure (SSIM).  The validation experiments show that the method can handle data from different trajectories, demonstrating its flexibility and robustness across different scan geometries. Our method demonstrates a significant improvement, particularly for the sinusoidal trajectory, achieving a 38.6\% reduction in MSE, a 7.7\% increase in PSNR, and a 5.0\% improvement in SSIM. Furthermore, the computation time for reconstruction was reduced by more than 97\%.
Code is available at \url{https://github.com/ChengzeYe/Defrise-and-Clack-reconstruction}.
\end{abstract}

%

\section{Introduction}
\label{sec:1}


Cone beam computed tomography (CBCT) is a widely used imaging technique in interventional medicine. It employs a cone-shaped X-ray beam, which captures a substantial amount of data in a single rotation, thereby facilitating the generation of three-dimensional (3D) volume while concurrently reducing radiation exposure.

Traditional CBCT scans usually involve a rotating structure with a fixed circular orbit for the X-ray source and detector. In contrast, recent robotic C-arm CT systems provide greater flexibility, which expand the scanning geometries beyond the conventional circular source-detector trajectory.

Flexible CT orbits offer numerous advantages, particularly in the field of interventional medicine. These advantages include enhanced image quality, an extended field of view (FOV), a reduced radiation dose \cite{schneider2022learning, schneider2023task}, and a reduction in metal artifacts \cite{herl2020scanning}.

However, if the orbit of the X-ray source follows a non-circular path, this presents a challenge to the reconstruction process. In such cases, the commonly used analytical algorithms, such as Filtered Back Projection (FBP), are no longer applicable, as they are based on the assumption of a circular trajectory.

Prior research has made significant strides in addressing the challenge of reconstructing images from non-circular orbital configurations.  For example, Zeng \cite{zeng2010medical} introduced an Algebraic Iterative Reconstruction (AIR) algorithm that is adept at reconstructing CBCT projections from arbitrary orbits.

A more sophisticated iterative algorithm, Model-Based Iterative Reconstruction (MBIR) \cite{liu2014model}, was subsequently developed. The application of MBIR has been demonstrated to markedly enhance image quality while simultaneously reducing noise and artifacts, thereby outperforming the AIR algorithm \cite{koetzier2023deep}. In addition to MBIR, some iterative reconstruction algorithms based on conjugate gradient descent are noted for their rapid convergence and low error, and can be adapted to arbitrary CBCT orbits~\cite{kawata1985constrained,liu2022improved}.

Due to the necessity of multiple iterations to progressively approximate the final reconstruction result, iterative algorithms have a high computational complexity and long processing times.

A notable contribution was made by Grangeat \cite{grangeat1991mathematical}, who proposed an exact reconstruction method for arbitrary orbits based on the relationship between the cone beam data and a function associated with the 3D Radon transform of the images. But this requires building an intermediate function matrix, which makes reconstruction relatively slow and memory intensive.



To address these limitations, Defrise and Clack \cite{defrise1994cone} introduced a filtered backprojection-type algorithm based on Grangeat's method. This approach customizes the algorithm for different orbits by designing specific redundancy weights, thereby achieving analytic reconstruction for any CBCT orbit. However, the analytic calculation of these weights for complex orbits in practical applications is a challenging task. 

In recent years, deep learning has become a widely used technique for fitting complex functions. But in the case of complex problems like the inverse problem in CT reconstruction, deep learning can lead to overparameterized models. Hence, Maier \textit{et al.} \cite{maier2019learning} proposed a method to address this issue, namely the integration of known operators as prior knowledge into machine learning models. This approach reduces the complexity of the model by minimizing the number of parameters, thereby enhancing the precision, interpretability and reliability of the model~\cite{sun2024data}. 

Building on the integration of known operators, our previous work provided an overview \cite{ye2024deep}  of the extension of the shift-variant FBP algorithm, leading to the development of a differentiable shift-variant FBP neural network for specific CBCT reconstruction. In this approach, the redundancy weights from Defrise and Clack's algorithm are optimized through a training process, where parameters are fitted based on a given circular orbit. This demonstrates the potential for rapid reconstruction of CBCT scans for any specific orbits.

This work builds upon our previous conference paper \cite{ye2024deep} propose a differentiable shift-variant FBP neural network designed for arbitrary CBCT orbits reconstruction using known operator learning. In comparison to our previous work, we concentrate on optimising the operators within the reconstruction pipeline and on the design of a suitable synthetic dataset, which considerably enhances the model's accuracy and fidelity. Furthermore, we conduct more extensive experiments to evaluate the model's capability in handling data from more complex non-circular orbits, and we validate its performance using medical data, thus ensuring greater robustness and practical applicability.

The remainder of this paper is organized as follows: Section~\ref{sec:2} outlines the theoretical background of our method. Sections~\ref{sec:3} and~\ref{sec:4} present experiments and results using circular, sinusoidal, circle plus arc orbits, and random nearest neighbor orbit. Finally, Section~\ref{sec:5} presents a discussion of the advantages and limitations of our model, while Section~\ref{sec:6} concludes the paper and outlines potential future directions for enhancing its performance.

\section{Methods}
\label{sec:2}

\subsection{Grangeat's Inversion}

Grangeat \cite{grangeat1991mathematical} proposed a reconstruction method that establishes a link between cone-beam projections and the first derivative of the Radon transform by defining an intermediate function. This method facilitates the rebinning process from the Cartesian coordinate system of cone-beam geometry to the spherical coordinate system of the Radon domain, utilising a plane as an information vector. Building upon this, Zeng \cite{zeng2010medical} outlines a process whereby a multitude of lines are sampled on each cone-beam projection. The line integral along these lines can be viewed as a plane integral weighted by $\frac{1}{r}$. The aforementioned plane is characterised by two parameters: the distance $l$ to the origin, which lies within the range $[-B, +B]$, and a vector, designated as $\theta$, which is orthogonal to the plane.

Based on this observation, Grangeat's intermediate function $S(\theta, \lambda )$~\eqref{eq:myequation2} can be expressed as follows:
\begin{align*}
S(\theta, \lambda )=-\int_{s^2}^{} d\beta \delta'(\beta, \theta)g(\beta,\lambda)\qquad\theta\in S^2, \lambda\in \Lambda, \tag{1}
\label{eq:myequation2}
\end{align*}
where $\lambda\in \Lambda$ denotes the parameter for the source position $a(\lambda)$, $\beta\in S^2$ (where $S^2$ is the set of all unit vectors in $\mathbb{R}^3$) indicates the direction of the line integral, the function $g(\beta,\lambda)$ represents the cone-beam projection, and $\delta'$ signifies the derivative of the Dirac delta distribution. 

The intermediate function can be incorporated into the inverse Radon transform formula, thereby deriving the reconstruction formula~\eqref{eq:myequation3000} for Grangeat's method:

\begin{align*}
f(x)=-\frac{1}{4\pi^2}\int_{s^2/2}^{}d\theta\int_{-B}^{B}dl \delta'(x\cdot\theta-l)S(\theta, \lambda(l,\theta) ).\tag{2}
\label{eq:myequation3000}
\end{align*}

This formular can already be adapted to arbitrary CBCT orbits. However, its main drawback is that it is not a FBP algorithm, as it necessitates data rebinning. The rebinning process entails interpolation, which may result in significant errors.

\subsection{Shift-variant filtered backprojection algorithm}
\label{sec:2.3}

Defrise and Clack \cite{defrise1994cone} introduced a shift-variant FBP algorithm based on Grangeat's method, designed for the reconstruction of projection data from general non-circular orbits. In contrast to Grangeat's algorithm, this approach circumvents the necessity to store intermediate functions on a discrete grid, thereby enabling the acquisition and reconstruction of projections to occur concurrently. This markedly accelerates the reconstruction process. The reconstruction formula is as follows:

\begin{align*}
f(x)=&\int_{ \Lambda}^{} d\lambda\int_{S^2/2}^{}d\theta-\frac{1}{4\pi^2}\mid a'(\lambda)\cdot\theta\mid\frac{1}{n(\theta, \lambda)}\\
&\times\delta' ((x-a(\lambda))\cdot \theta)S(\theta, \lambda ),\tag{3}
\label{eq:myequation3}
\end{align*}
where the term $n(\theta, \lambda)$ is defined as the number of intersections between orbit $a(\Lambda)$ and the plane orthogonal to $\theta$ through orbit point $a(\lambda)$. 


Equation~\eqref{eq:myequation3} can be broken down into three distinct components. Firstly, for each cone-beam projection $\lambda\in \Lambda$, Grangeat's intermediate function is computed using~\eqref{eq:myequation2}. Second, shift-variant filtering is applied to the resulting intermediate function:
\begin{align*}
g^F(\omega, \lambda)=& \int_{\substack{S^2/2}} d\theta-\frac{1}{4 \pi^2} \mid a'(\lambda)\cdot\theta\mid\frac{1}{n(\theta, \lambda)}\delta'(\omega, \theta)S(\theta,\lambda) \\
&\omega\in S^2, \lambda\in \Lambda.\tag{4}
\label{eq:myequation4}
\end{align*}

Finally, as with the conventional Feldkamp-Davis-Kress (FDK) algorithm, the filtered cone-beam projections are backprojected into the 3D volume:
\begin{align*}
f(x)=&\int_{\substack{\Lambda}} d\lambda\frac{1}{\mid x-a(\lambda)\mid^2}g^F\left(\frac{x-a(\lambda)}{\mid x-a(\lambda)\mid}, \lambda\right) \\
&x\in \mathbb{R}^3, \mid x\mid\leq B.\tag{5}
\label{eq:myequation5}
\end{align*}

\subsection{Differentiable shift-variant FBP neural network}
\label{sec:2.4}
In our preceding work \cite{ye2024deep}, we gave an overview of how Defrise and Clack's algorithm into a trainable neural network. In the following, we summarize the most important steps to derive the differentiable formulation of the Defrise and Clack's algorithm.

\begin{figure}[!t]
\centering
\includegraphics[width=2.5in]{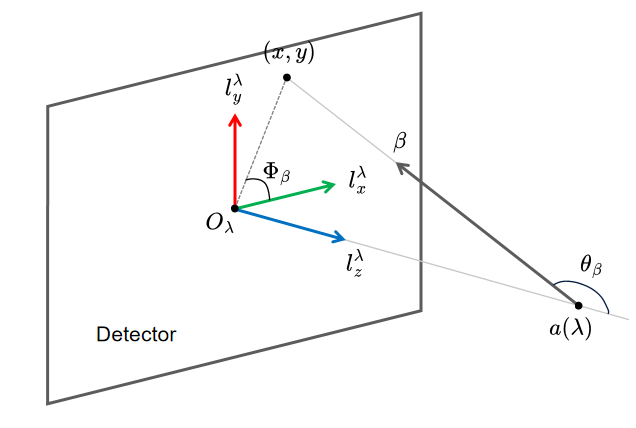}
\caption{Detector coordinate system with origin $O_\lambda$ at the centre of the detector. The vectors $l_x^\lambda$ and $l_y^\lambda$ lying in the detector plane, while $l_z^\lambda$ as the unit vector pointing towards the source}
\label{detector coordinate system}
\end{figure}

For each projection, it is first necessary to compute Grangeat's intermediate function $S(\theta,\lambda)$, which serves to transform the cone-beam projection $g(x,y,\lambda)$ from the detector coordinate system into the Radon domain. As a consequence, it is first necessary to define a detector coordinate system for each detector position, as illustrated in Figure~\ref{detector coordinate system}.

Furthermore, it is necessary to establish a sinogram coordinate system. The function $S(\theta,\lambda)$ is dependent upon the unit vector $\theta$, which defines a plane that is orthogonal to $\theta$ and passes through the source. This plane intersects the detector plane along a line, which can be parameterised by the perpendicular distance $s$ from the detector centre to the line and the angle $\mu$ between this perpendicular and the x-axis.

In accordance with the definitions provided, Equation~\eqref{eq:myequation11} can be derived from ~\eqref{eq:myequation2} as presented by Defrise and Clack~\cite{defrise1994cone}:

\begin{align*}
S(s,\mu,\lambda)=\frac{s^2+D^2}{D^2}\int_{+e}^{-e}dv\frac{\partial }{\partial u}\left\{\frac{Dg(x,y,\lambda)}{\sqrt{u^2+v^2+D^2}}\right\} _{u=s}.\tag{6}
\label{eq:myequation11}
\end{align*}

In this equation, $u$ and $v$ represent coordinate transformations defined as $u=x\cos\mu+y\sin \mu$ and $v=-x\sin\mu+y\cos \mu$, while $D$ denotes the distance between the source and the detector. In addition, the variable $e$ represents the radius of the region defined by the cone-beam projection of the field of view.

According to~\eqref{eq:myequation11}, The calculation of Grangeat's intermediate function can be divided into three stages, each of which corresponds to a layer in the neural network architecture. The initial stage of the process entails the application of a cosine weighting function $w_{cos}$, defined as $w_{cos}=\frac{D}{\sqrt{u^2+v^2+D^2}}$, to the input projection data $g(x,y,\lambda)$. 
In the second step, a 2D Radon transform $A_{2d}$ is performed using parallel beam geometry to project the weighted cone-beam data. This step yields the representation of the weighted cone-beam projection in the Radon domain. Finally, differentiation $D$ with respect to $s$ of the resulting sinogram is computed. Following differentiation, an additional weighting function $w_{sino}$, defined as $w_{sino}=\frac{s^2+D^2}{D^2}$, is applied within the sinogram domain. 



Collectively, these operations constitute the computational pipeline for generating Grangeat's intermediate function in the neural network framework, ultimately leading to as follows:
\begin{align*}
S(s,\mu,\lambda)=w_{sino}DA_{2d}w_{cos}g(x,y,\lambda).\tag{7}
\label{eq:myequation12}
\end{align*}

\begin{figure}[!t]
\centering
\includegraphics[width=6in]{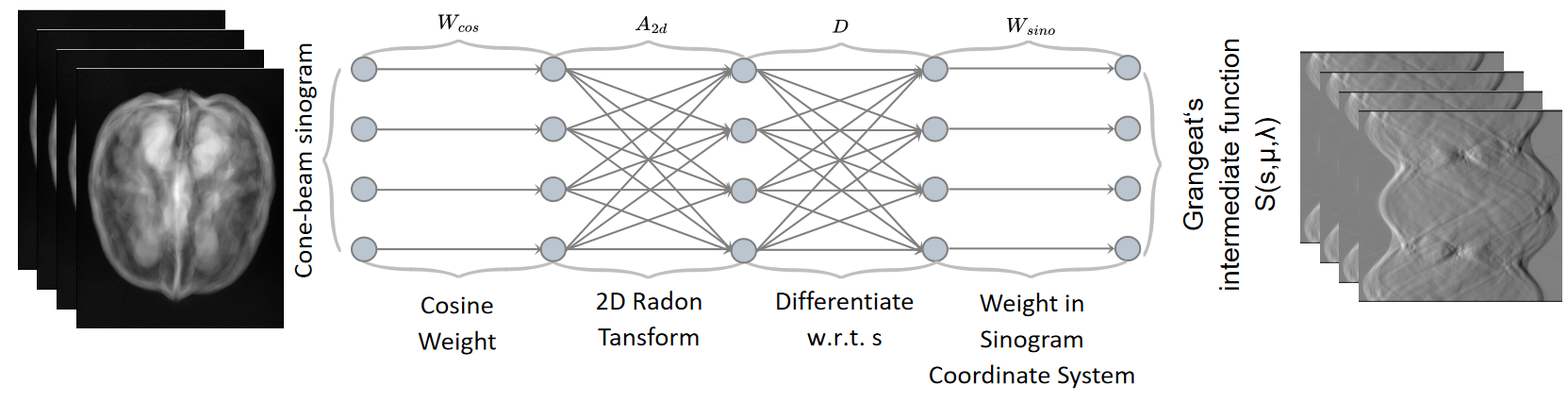}
\caption{Grangeat’s intermediate function part of the neural network: The input of the neural network is cone-beam projections, and the output is grangeat’s intermediate function.}
\label{network1}
\end{figure}

This equation differs from~\eqref{eq:myequation11} mainly in the order of integration and differentiation being switched. The purpose of this switch is to reduce the noise caused by the differentiation process. Using~\eqref{eq:myequation12}, we construct the neural network shown in Figure \ref{network1}

In accordance with the description provided in Section~\ref{sec:2.4}, the computed intermediate function must be processed with a shift-variant filter in order to obtain the filtered cone-beam projection. Specifically, it is necessary to explicitly rewrite~\eqref{eq:myequation4} for the case where the cone-beam data are acquired on a planar detector. This process leads to the derivation of~\eqref{eq:myequation13}, as originally presented by Defrise and Clack \cite{defrise1994cone}:

\begin{align*}
g^F(x,y,\lambda)&=(x^2+y^2+D^2)\int_{0}^{\pi} d\mu\left\{\frac{\partial }{\partial s}\frac{S_1(s,\mu,\lambda)}{\sqrt{s^2+D^2}}\right\} \\
s&=x\cos\mu+y\sin\mu \\ 
S_1(s,\mu,\lambda)&=-\frac{1}{4\pi^2}\mid a'(\lambda)\cdot\theta\mid M(\theta,\lambda)S(\theta,\lambda).\tag{8}
\label{eq:myequation13}
\end{align*}


In the original equation~\eqref{eq:myequation4}, the function $n(\theta, \lambda)$ is discontinuous, and this discontinuity can lead to the formation of artifacts in a discrete implementation. Defrise and Clack addressed this issue by substituting the crofton symbol $\frac{1}{n(\theta, \lambda)}$ with a differentiable continuous function:
\begin{align*}
M(\theta,\lambda) = \frac{\mid a'(\lambda)\cdot\theta\mid^m c(\lambda)}{\sum_{\alpha=1}^{n(\theta, \lambda)} \mid a'(\lambda_\alpha)\cdot\theta\mid^m c(\lambda_\alpha)
}, \tag{9}
\label{eq:myequation111}
\end{align*}
where $m$ be a positive integer, $c(\lambda)$ is a smooth function that is equal to one throughout the entire interval $\Lambda$, with the exception of the region near the interval boundaries. 

It is of great importance to obtain a continuous, differentiable function with properties similar to the crofton symbol in order to achieve optimal reconstruction quality. However, the practical application of Eq.~\eqref{eq:myequation111} is subject to numerous constraints. For instance, how to appropriately choose the parameter $m$ based on the complexity of the orbit geometry. if the constant $m$ is not learned in a data-driven manner, it may not represent the optimal choice for different trajectories and applications.

In this work, we refer to the product of orbit-related term $\mid a'(\lambda)\cdot\theta\mid$ and the crofton symbol $\frac{1}{n(\theta, \lambda)}$ as the redundancy weight. This weight can be treated as trainable parameters in the reconstruction pipeline, and automatically estimated based on the given orbital geometry through the training of the neural network.

Similarly, Equation~\eqref{eq:myequation13} can be decomposed into four steps, with each step corresponding to a layer in the neural network. First, the redundancy weight, defined as follows:
\begin{align*}
w_{red}(s,\theta,\lambda)=-\frac{1}{4\pi^2} M(\theta,\lambda)\frac{\mid a'(\lambda)\cdot\theta\mid}{\sqrt{s^2+D^2}}, \tag{10}
\label{eq:myequationx3}
\end{align*}
is applied to the intermediate function. This corresponds to the redundancy weight layer in the network, which is the only layer containing trainable parameters. Subsequently, the weighted intermediate function is differentiated with respect to $s$ in the sinogram domain using the operator $D$. Then, the sinogram after differentiation is backprojected into the detector coordinate system using the same parallel beam geometry via the backprojection operator $A_{2d}^T$. Ultimately, the weight $w_d=x^2+y^2+D^2$ is applied in the detector domain for geometric correction. By combining the above steps, the following formula is obtained:
\begin{align*}
g^F(x,y,\lambda)=w_{d}A_{2d}^TDw_{red}S(s,\mu,\lambda).\tag{11}
\label{eq:myequation14}
\end{align*}

\begin{figure}[!t]
\centering
\includegraphics[width=6in]{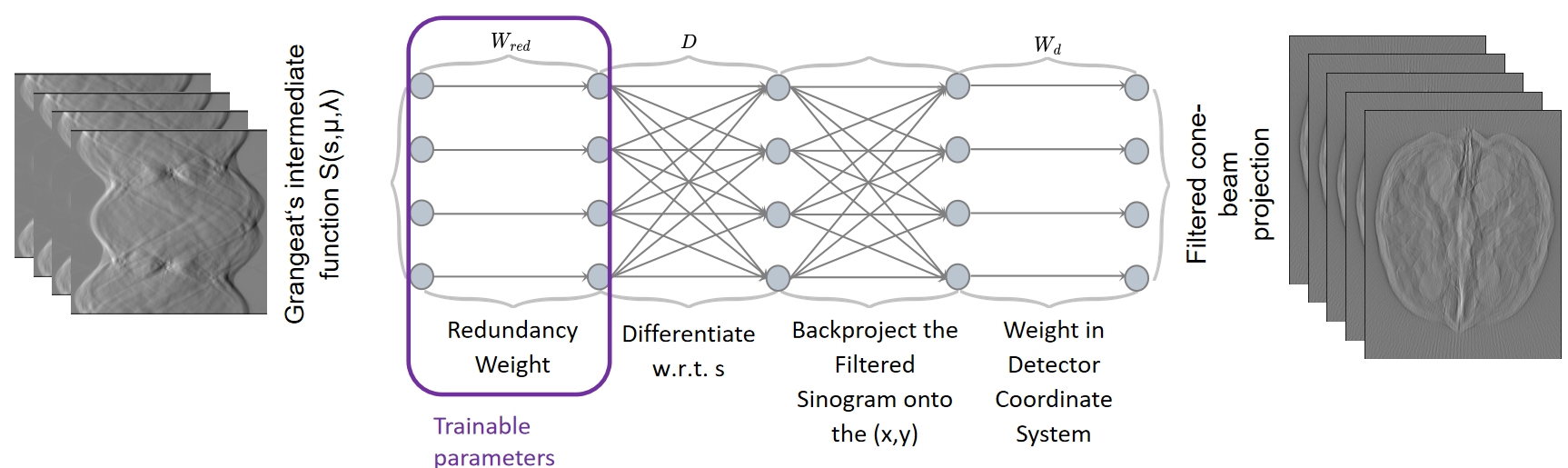}
\caption{Filtering part of the neural network: The input of the neural network is grangeat’s intermediate function, and the output is filtered cone beam projections.}
\label{network2}
\end{figure}

Based on eq.~\eqref{eq:myequation14}, the filtering step of the neural network can be constructed as shown in Figure \ref{network2}.

\begin{figure*}[!t]
\centering
\includegraphics[width=2.5in]{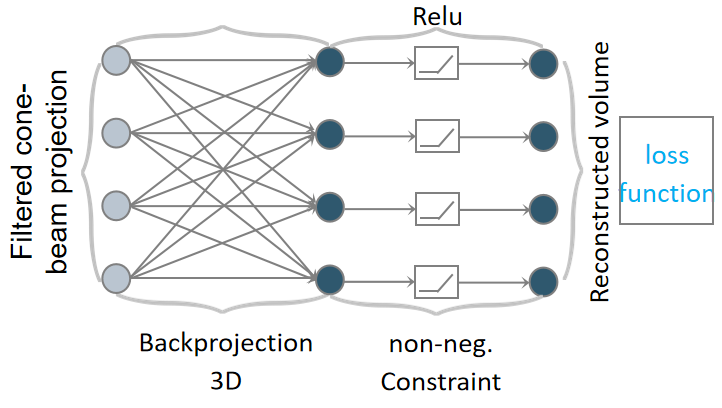}
\caption{Backprojection part for the neural network: Backproject the filtered cone-beam projections into 3D volume}
\label{network3}
\end{figure*}


The aforementioned steps involve shift-variant filtering of the cone-beam projection data, which is subsequently backprojected into a 3D volume, designated as $A_{3d}^T$, which serves as the ultimate layer of the network. Following the 3D back-projection, a ReLU activation function is applied to ensure non-negativity of the output, as illustrated in Figure~\ref{network3}.

By combining the backprojection to 3D with Grangeat’s intermediate function calculation~\eqref{eq:myequation12} and shift variant filtering~\eqref{eq:myequation14}, we obtain the complete reconstruction formula: 
\begin{align*}
f(x)=A_{3d}^Tw_{d}A_{2d}^TDw_{red}w_{sino}DA_{2d}w_{cos}g(x,y,\lambda).\tag{12}
\label{eq:myequation15}
\end{align*}

\section{Experiments}
\label{sec:3}
The differentiable shift-variant FBP neural network exhibits trajectory geometric specificity, meaning that when the input projection data corresponds to different geometric trajectories, the model must be trained for each trajectory geometry.  
In order to validate the performance of our model, we utilise three distinct types of trajectory geometries: circular, sinusoidal, circle plus arc, and random nearest neighbor orbit. We generate simulated datasets based on each of these geometries, which is then employed to train the neural network and optimise its parameters. 
Meanwhile, a test set following the same geometry generated from real data is used to evaluate the network's performance. 


\subsection{Implementation Details}
\begin{figure*}[b] 
\centering
\includegraphics[width=7in]{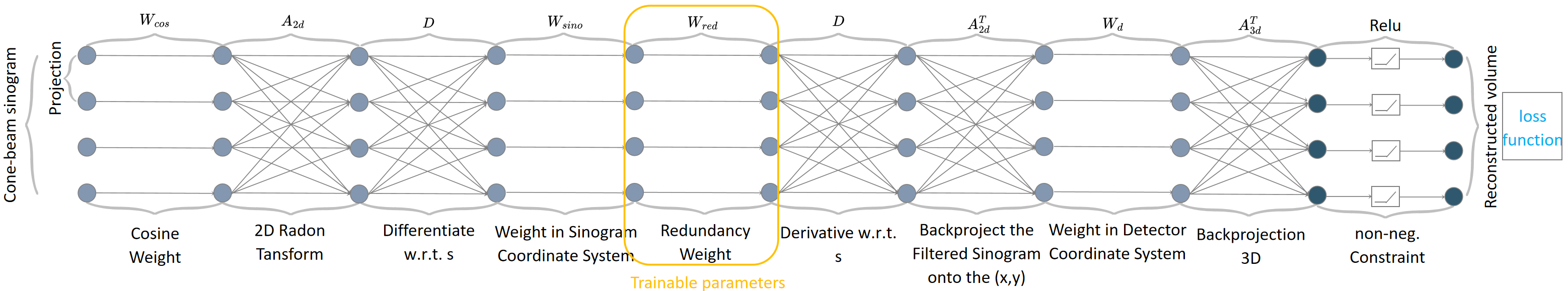}
\caption{Grangeat's intermediate function part, filtering part, and backprojection part are combined to form the differentiable shift-variant FBP neural network architecture.}
\label{network}
\end{figure*}

Based on ~\eqref{eq:myequation15}, the complete network architecture has been constructed, as shown in Figure \ref{network}. 

To ensure smoothness in the learned redundancy weights, a Gaussian filter layer with a sigma value of 20 and a kernel size of 121 was added after the redundancy weight layer. This effectively reduces the noise interference in reconstruction process and improves the reconstruction quality.

We used PyTorch 2.1.1 to construct the neural network's architecture. The 2D Radon transform and the 3D cone-beam backprojection were implemented using operators from the framework PyroNN \cite{syben2019pyro}. 

The loss function for training is formulated as:
\begin{align*}
\mathcal{L}_{total} = \mathcal{L}_{mse} + \gamma \cdot\mathcal{L}_{ssim},\tag{13}
\label{eq:myequation18}
\end{align*}
where $\mathcal{L}_{mse}$ represents the mean squared error (MSE) loss, $\mathcal{L}_{ssim}$ denotes structural similarity index (SSIM) loss, and $\gamma$ is a weighting factor. Following the normalization of the reconstructed result and the reference, the loss was calculated and the parameters were updated using the AdamW optimiser. The application of the One-cycle learning rate policy \cite{smith2018disciplined}, which adjusted the learning rate cyclically between $0.1$ and $1$ over $1000$ epochs, resulted in a notable enhancement in the training performance. Furthermore, a random initialisation was conducted for the redundancy weight layer $W_{red}$, utilising values that were uniformly distributed between -1 and 0. Finally, the training process was conducted on an Nvidia A40 GPU.


\subsection{Geometry Configuration}

In our experiments, we employ the precise geometry parameters of a clinical C-arm system, the Artis zeego (Siemens AG, Forchheim, Germany). The detector matrix is operated in a 4 × 4 binning mode with 620 × 480 pixels at an isotropic pixel size of 0.616 mm. The source-to-detector distance is 1200 mm, and the source-to-center of rotation distance is 750 mm. A total of 400 projections are utilized for reconstruction. The reconstruction matrix is of dimensions 128 × 512 × 512, with a voxel size of 0.25 × 0.25 × 0.25 mm.
In distinct experiments, the positions of the sources follow circular, circle plus arc, sinusoidal trajectories, and random nearest neighbor orbit, respectively.

The aforementioned parameters are presented in Table \ref{tab: Geometry Configuration} for convenient reference.

\begin{table}[htbp]
\centering
\small
\caption{Geometry configuration}
\begin{tabular}{{@{}cccc@{}}}
\toprule
\textbf{Volume shape} & \textbf{Volume spacing} & \textbf{Number of projections} \\ 
\midrule
128$\times$512$\times$512 & 0.25mm$\times$0.25mm$\times$0.25mm & 400 \\

\toprule
\textbf{Source isocenter distance} & \textbf{Source detector distance}& \textbf{Detector spacing}\\
\midrule
 750mm & 1200mm& 0.616mm$\times$0.616mm \\
 
\toprule
\textbf{Detector shape} &\textbf{Source geometry}   \\ 
\midrule
620$\times$480&Cone beam  \\
\bottomrule
\end{tabular}%
\label{tab: Geometry Configuration}
\end{table}

Table \ref{tab: Geometry configuration for detector sampling} presents the geometric parameters utilized in the 2D Radon transform and backprojection processes within the differentiable shift-variant FBP reconstruction pipeline. The parameters are selected based on the detector shape and spacing to ensure complete sampling of the projection data.

\begin{table}[htbp]
\centering
\small
\caption{Parallel beam geometry configuration for detector sampling}
\begin{tabular}{{@{}cccc@{}}}
\toprule
\textbf{Volume shape} & \textbf{Volume spacing} & \textbf{Number of projections} \\ 
\midrule
620$\times$480 & 0.616mm$\times$0.616mm & 360 \\

\toprule
\textbf{Angular range} & \textbf{Source isocenter distance} & \textbf{Source detector distance}\\
\midrule
$\pi$ & 241mm & 483mm \\
 
\toprule
\textbf{Detector shape} & \textbf{Detector spacing}  \\
\midrule
785 & 0.616mm\\
\bottomrule
\end{tabular}%
\label{tab: Geometry configuration for detector sampling}
\end{table}

\subsection{Training Data}
\label{sec:Training data}

A total of $30$ simulated data samples are generated in order to create a suitable training and validation dataset. $24$ of the samples are utilized for training, while the remaining samples are employed for validation. 

\begin{figure}[!t]
\centering
\includegraphics[width=2.5in]{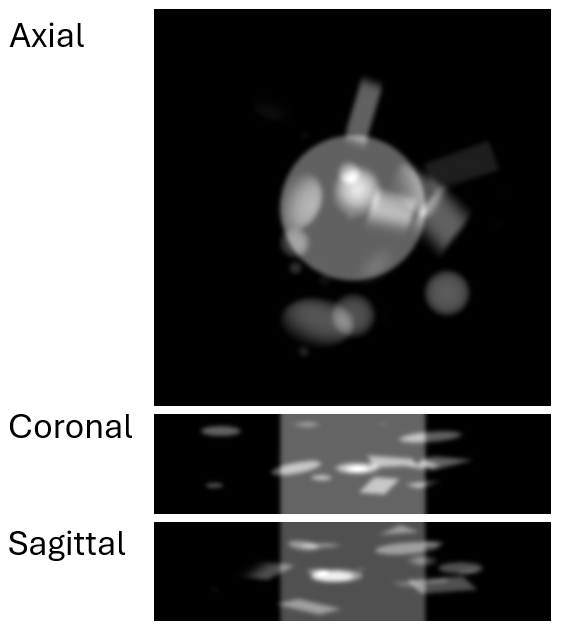}
\caption{Central slices of the ground truth.}
\label{GT}
\end{figure}

As the ground truth in each sample, a random number of geometric objects are generated within a voxel volume. These objects vary in type, position, and rotation direction. Furthermore, we incorporate cylinders with randomly generated diameters along the first dimension to make the simulated dataset more similar to the human thoracic and abdominal structures. Finally, in order to mitigate the impact of sharp edges on neural network training, Gaussian filtering is applied. Figure \ref{GT} illustrates central slices of the ground truth obtained from three directions.

We follow the orbit geometry presented in Table \ref{tab: Geometry Configuration} and use the cone beam forward projection of the software PyroNN \cite{syben2019pyro} to generate the sinogram, which serves as input for the neural network.

\subsection{Real Data}
In order to assess the efficacy of the differentiable shift-variant FBP neural network, we utilized the Pancreatic-CT-CBCT-SEG dataset \cite{hong2021breath}, which was provided by the Memorial Sloan Kettering Cancer Center and included CT imaging data from 40 patients with pancreatic cancer. 

A total of five samples were selected in the experiment, using the same approach employed in Section  \ref{sec:Training data} to generate the corresponding sinogram for testing purposes.

\section{Results}
\label{sec:4}

\subsection{Circular Orbit}
\label{sec:4.1}

First, we employ the circular orbit data, as its straightforward structure permits the straightforward derivation of the analytical solution for comparison.

The analytical solution of the redundancy weight for circular orbit is defined as follows:
\begin{align*}
M(\theta,\lambda) =& \frac{1}{2} \\
\mid a'(\lambda)\cdot\theta\mid=&\frac{D^2 \mid \cos\mu \mid }{\sqrt{s^2+D^2}}, \tag{14}
\label{eq:myequation16}
\end{align*}
which reveals that each source position corresponds to an identical redundancy weight. Leveraging this insight as prior knowledge can significantly reduce the number of parameters required by the differentiable shift-variant FBP neural network in such scenarios, thereby accelerating convergence rates. Specifically, the network converges within $50$ epochs when the learning rate is set to $0.001$.

\begin{figure}[!t]
    \centering
    \subfloat[]{\includegraphics[width=2.5in]{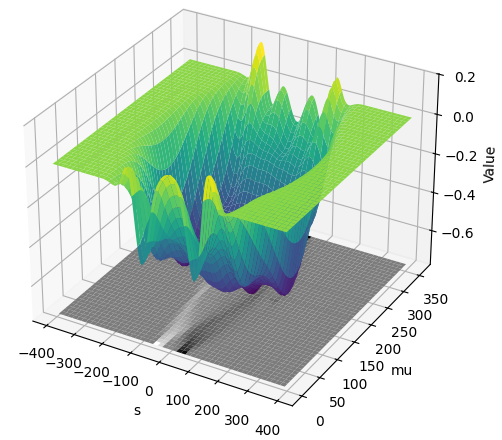}}
    \subfloat[]{\includegraphics[width=2.5in]{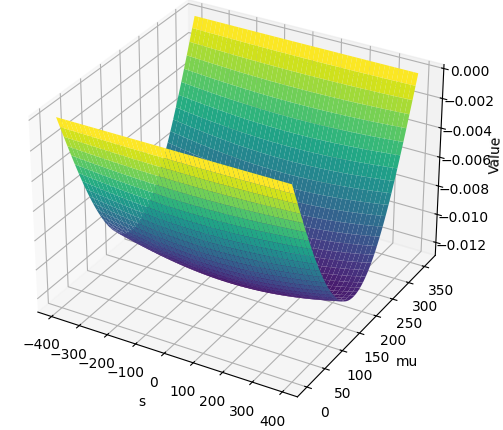}}
    \caption{Learned redundancy weights (circular orbit). (a) Learned weight. (b) Analytic weight.}
    \label{learned_weight}
\end{figure}

In Figure \ref{learned_weight}, the visualisation of learned redundancy weight layer parameters is presented, compared with the analytical solution, and it is demonstrated that they have similar structures. However, the redundancy weights display distinct characteristics when compared to the analytical solution in regions where the value of $s$ is either extremely large or small. This is due to the fact that the Grangeat’s intermediate function is equal to zero in these regions, which results in a zero contribution to the gradient update of the weights.

Table \ref{tab: metricscircular} employs MSE, Peak Signal-to-Noise ratio(PSNR), and SSIM as quantitative evaluation metrics to analyze the efficacy of the learned weights for reconstruction. 
Given that the reconstructed volume exhibits discrepancies in pixel intensity when compared to the ground truth, we apply histogram matching to the reconstructed result prior to calculating the quantitative metrics.
It can be observed that the learned redundancy weights yield results that are in close alignment with the analytic redundancy weights, with only minor discrepancies. These minor discrepancies may be attributed to the selection of the loss function or the optimiser's hyperparameters.

\begin{figure}[!t]
\centering
\includegraphics[width=5.5in]{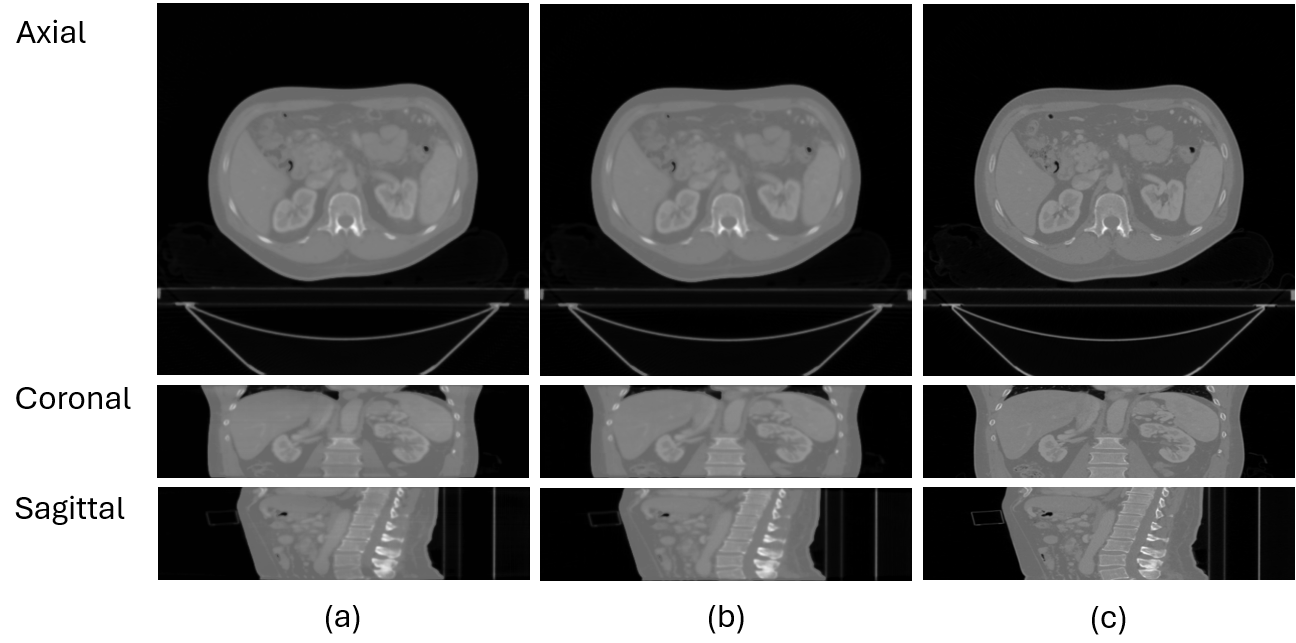}
\caption{Reconstructed results of the network (Circular orbit). (a) Reconstruction using learned redundancy weight. (b) Reconstruction using analytic redundancy weight. (c) Ground truth.}
\label{result(circular)}
\end{figure}

\begin{table}[htbp]
\centering
\small
\caption{Comparison of Image Quality Metrics (Circular orbit) with Mean ± Standard Deviation}
\begin{tabular}{@{}lccc@{}}
\toprule
\textbf{Weighting type} & \textbf{MSE}$\downarrow$ & \textbf{PSNR (dB)}$\uparrow$ & \textbf{SSIM}$\uparrow$\\ 
\midrule
Learned & 0.0945± 0.0136&  36.79±1.38 & 0.9531±0.0057\\
Analytical & 0.0983±0.0126  &  36.43±1.44 & 0.9684±0.0036\\
\bottomrule
\end{tabular}%
\label{tab: metricscircular}
\end{table}
In Figure \ref{result(circular)}, a comparison of the central slice of volumes reconstructed with different redundancy weights is presented. The results show that the neural network, after training, can achieve reconstruction results similar to those obtained through the analytical method.

\subsection{Sinusoidal Orbit}
\label{sec:4.2}

\begin{figure}[!t]
\centering
\includegraphics[width=3.6in]{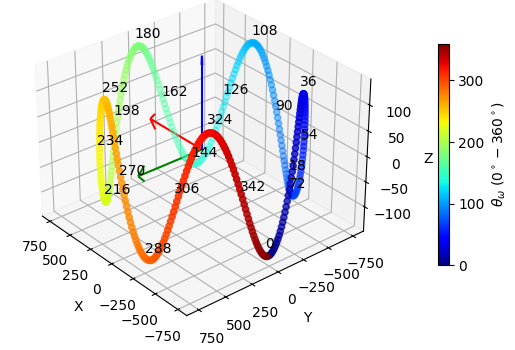}
\caption{Sinusoidal orbit.}
\label{Sinusoidal Orbit}
\end{figure}

The sinusoidal trajectory is a commonly used trajectory in the field of medical imaging. It has been demonstrated that this trajectory can effectively reduce metal artifacts \cite{gang2020metal}. In the spherical coordinate system, it can be expressed as follows: 
\begin{align*}
  \phi_\omega &= \phi_{max} \cos(f \theta_\omega),\tag{15}
\label{eq:myequation20}
\end{align*}
where $\theta_\omega$ represents the traditional gantry rotation angle, which ranges from $-180^\circ$ to $180^\circ$. Additionally, $\phi_\omega$ denotes the gantry tilt angle. The maximum tilt angle is represented by $\phi_{max} $. $f$ defines the frequency of the sinusoid. In this experiment, a sinusoidal trajectory with a maximum tilt angle of $20^\circ$ and a frequency of $5$ is employed, as illustrated in Figure~\ref{Sinusoidal Orbit}. 

\begin{figure}[!t]
    \centering
    \subfloat[]{\includegraphics[width=1.5in]{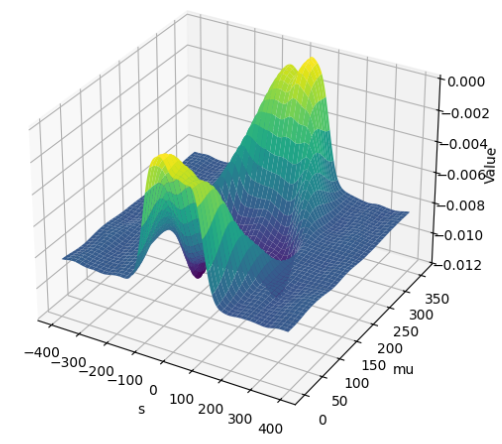}}
    \subfloat[]{\includegraphics[width=1.5in]{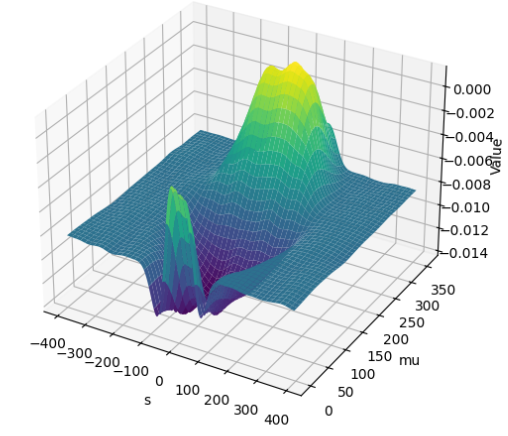}}
    \subfloat[]{\includegraphics[width=1.5in]{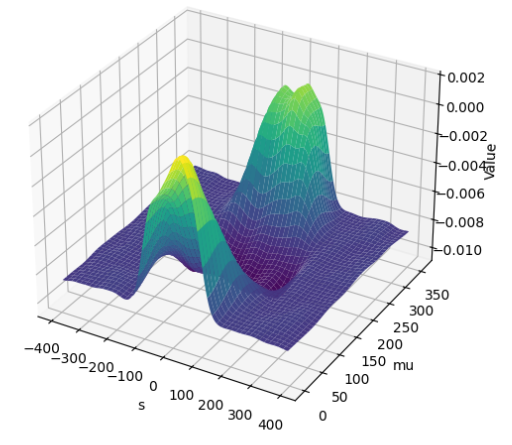}}
    \subfloat[]{\includegraphics[width=1.5in]{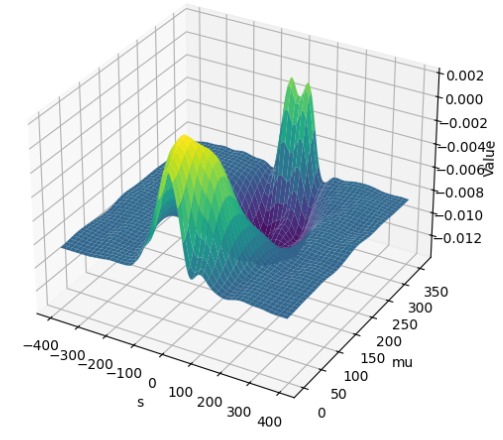}}
    \caption{Learned redundancy weights (Sinusoidal Orbit). (a)$\theta_\omega = 0^\circ$. (b) $\theta_\omega = 18^\circ$. (c)$\theta_\omega = 36^\circ$. (d) $\theta_\omega = 54^\circ$.}
    \label{learned_weight(Sinusoidal Orbit)}
\end{figure}

Figure~\ref{learned_weight(Sinusoidal Orbit)} depicts the parameters within the redundancy weight layer following the convergence of the neural network after $430$ epochs with an initial learning rate of $0.1$. It can be observed that as the trajectory undergoes periodic alterations, the redundancy weight also exhibits corresponding periodic behaviors.

Table \ref{tab: Sinusoidal orbit} illustrates that, in comparison to conventional AIR algorithms, our methodology demonstrates superior performance across a range of metrics, including MSE, SSIM, and PSNR. In particular, the method achieves a lower MSE, indicating a higher level of accuracy, and shows improved SSIM and PSNR, reflecting an enhanced fidelity and image quality.~Moreover, the times indicated in the table represent the reconstruction duration on an NVIDIA 4080 Super GPU. This demonstrates that our approach after training significantly outperforms conventional AIR algorithm in terms of reconstruction speed.

\begin{table}[htbp]
\centering
\small
\caption{Comparison of Image Quality Metrics (Sinusoidal orbit) with Mean ± Standard Deviation}
\begin{tabular}{@{}lcccccc@{}}
\toprule
 & \textbf{MSE}$\downarrow$ & \textbf{PSNR (dB)}$\uparrow$ & \textbf{SSIM}$\uparrow$& \textbf{Time (s)}$\downarrow$\\ 
\midrule
Our & 0.0904 ± 0.0149&  37.20± 1.34& 0.9591±0.0051 & 4.5\\
AIR(300) & 0.1472± 0.0827&  34.55± 4.11& 0.9131±0.0527 & 197\\
\bottomrule
\end{tabular}%
\label{tab: Sinusoidal orbit}
\end{table}

\begin{figure}[!t]
\centering
\includegraphics[width=5.5in]{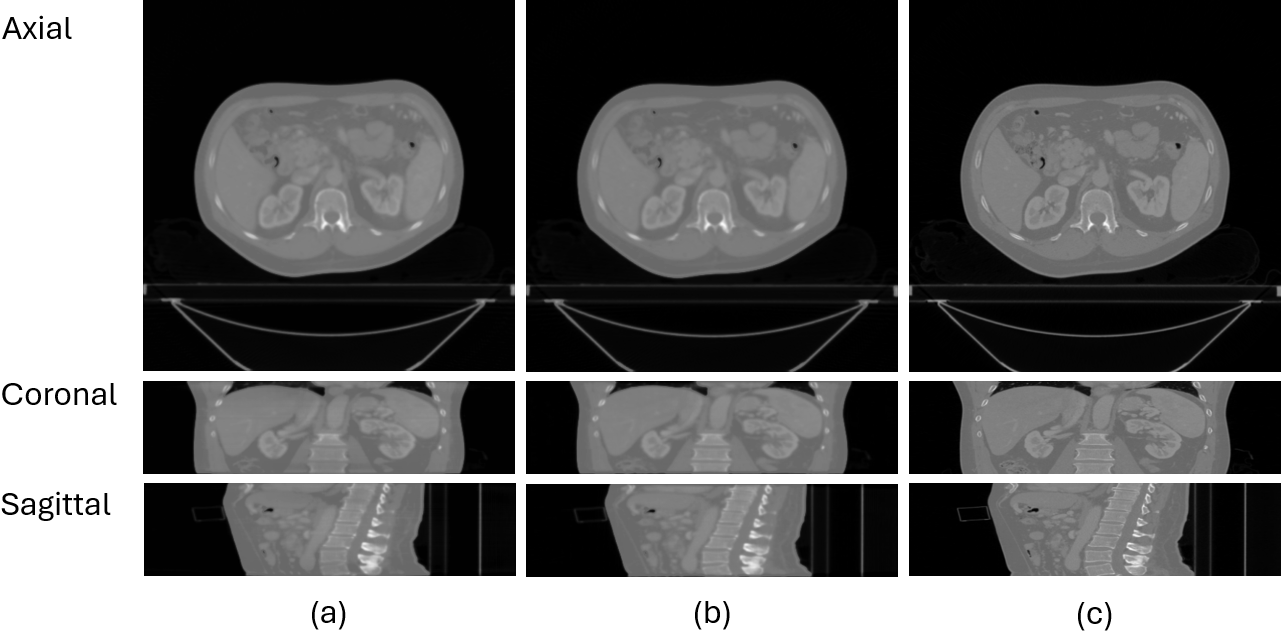}
\caption{Reconstructed results for the network(Sinusoidal orbit). (a) Reconstruction using learned weights. (b) Iterative reconstruction result (300 iterations). (c) Ground truth.}
\label{result(Sinu)}
\end{figure}

Figure \ref{result(Sinu)} presents the reconstructed volume obtained with learned parameters.
A comparison of the results obtained from the AIR algorithm with the ground truth demonstrates that the differentiable shift-variant FBP model is an effective method for preserving reconstruction details while maintaining structural accuracy.

\subsection{Circle Plus Arc Orbit}
\begin{figure}[!t]
\centering
\includegraphics[width=3.5in]{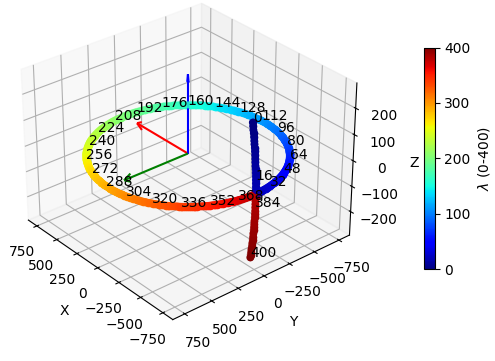}
\caption{Circle Plus Arc Orbit. The numbered labels in the figure, denoted by $\lambda$, are in accordance with the sequence of projections, ranging from $0$ to $399$, which represent a total of $400$ projections.}
\label{CirclePlusArc}
\end{figure}

Another common non-circular orbit is the "Circle Plus Arc Orbit", as illustrated in Figure \ref{CirclePlusArc}. This orbit combines a circle path with a small orthogonal arc. This orbit fulfills Tuy's data sufficiency condition, effectively addressing the issue of decreased reconstruction accuracy at locations distant from the central transverse plane when the cone angle is considerable. 

\begin{figure}[!t]
    \centering
    \subfloat[]{\includegraphics[width=2.5in]{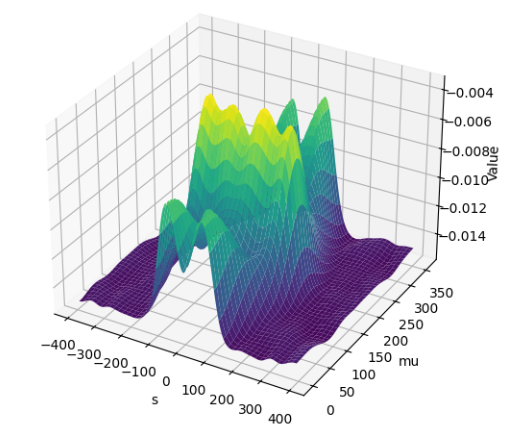}}
    \subfloat[]{\includegraphics[width=2.5in]{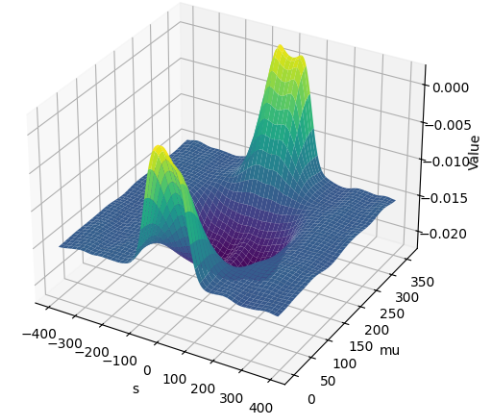}}
    \caption{Learned redundancy weights (Circle Plus Arc Orbit). (a) $\lambda = 10$ (on the arc). (b) $\lambda = 60$ (on the circle).}
    \label{learned_weight(Circle Plus Arc Orbit)}
\end{figure}

\begin{figure}[!t]
\centering
\includegraphics[width=5.5in]{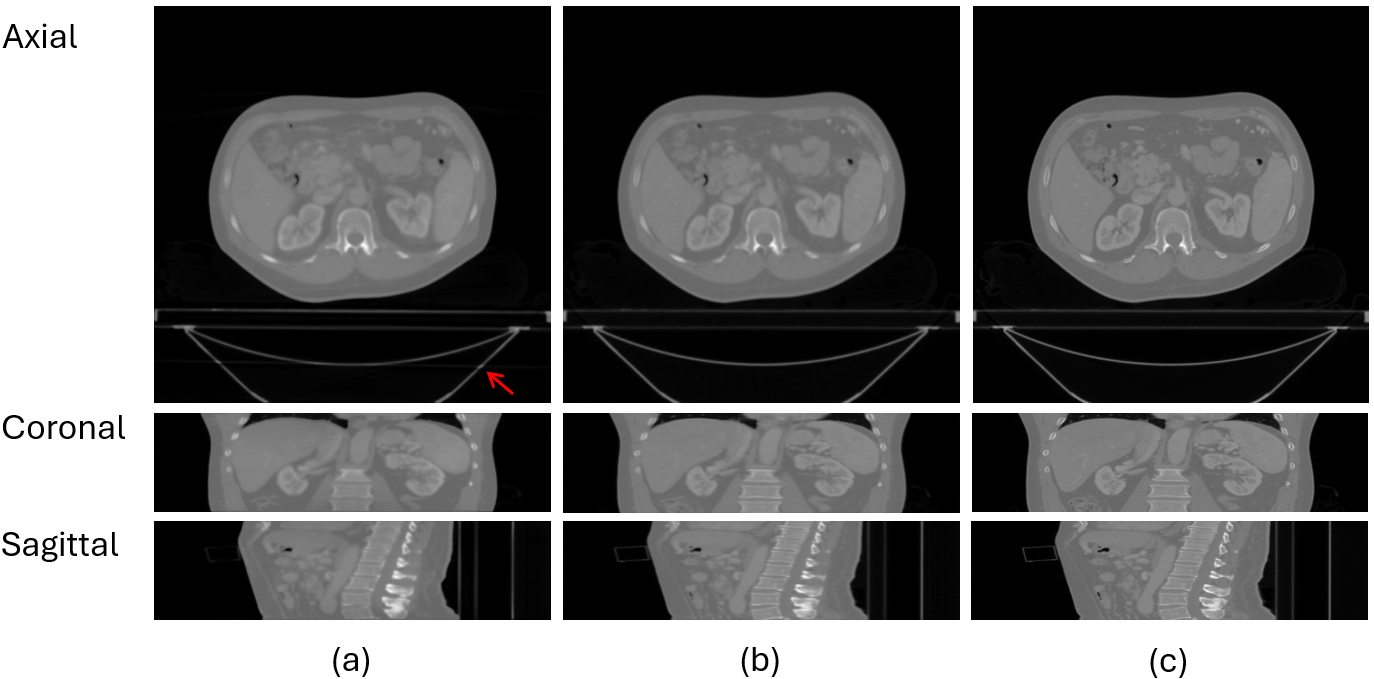}
\caption{Reconstructed results for the network(Circle Plus Arc Orbit). (a) Reconstruction using learned weights. (b) Iterative reconstruction result (300 iterations). (c) Ground truth.}
\label{result(arc)}
\end{figure}

A learning rate of $0.1$ and training for $400$ epochs resulted in successful convergence of the network. Figure~\ref{learned_weight(Circle Plus Arc Orbit)} depicts the learned redundancy weights obtained through the training process. It is evident that the projection data requires a fundamentally different redundancy weight when the source is on the circle than when the source is on the arc. These learned weights were then employed for reconstruction and compared with the results from iterative reconstruction and the ground truth. The reconstruction results, visualised using central slices, are presented in Figure~\ref{result(arc)}.

Although the overall reconstruction quality is satisfactory, it is evident that there are some artifacts present at the location indicated by the red arrow in the central slice of the axial direction. Additionally, there is slight blurring visible in some regions. The aforementioned artifacts and blurring are caused by the discontinuities that occur at the endpoints of the trajectory, given that the trajectory is not closed.

As in Sections \ref{sec:4.1} and \ref{sec:4.2}, the quality of the reconstructed volumes was assessed using MSE, PSNR, and SSIM as shown in Table~\ref{tab: Circle Plus Arc Orbit}. The results demonstrate that all evaluation metrics performed exceptionally well, indicating that suitable redundancy weights can be effectively learned based on the Circle Plus Arc Orbit.

\begin{table}[htbp]
\centering
\small
\caption{Comparison of Image Quality Metrics (Circle Plus Arc Orbit) with Mean ± Standard Deviation}
\begin{tabular}{@{}lcccccc@{}}
\toprule
 & \textbf{MSE}$\downarrow$ & \textbf{PSNR (dB)}$\uparrow$ & \textbf{SSIM}$\uparrow$& \textbf{Time (s)}$\downarrow$\\ 
\midrule
Our & 0.1075 ± 0.0185&  35.71± 1.49& 0.9339±0.0055 & 4.5\\
AIR(300) & 0.1360± 0.0565&  34.41± 2.53& 0.9216±0.0365 & 190\\
\bottomrule
\end{tabular}%
\label{tab: Circle Plus Arc Orbit}
\end{table}

\subsection{Random Nearest Neighbor Orbit}
\begin{figure}[!t]
\centering
\includegraphics[width=3.6in]{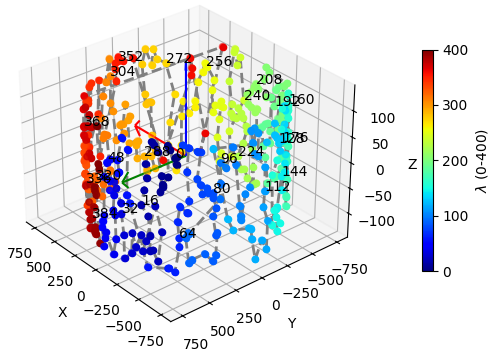}
\caption{RandomOrbit.}
\label{RandomOrbit}
\end{figure}

In order to further explore the potential of our model, we designed a geometric configuration with extremely complex trajectory. This trajectory is based on a spherical surface, with the assumption that the scanned object is located at the center of the sphere. 
As a consequence of the physical and geometric constraints inherent to the robotic C-arm CT system, the source points are constrained to a particular surface area near the middle plane of the sphere. Within this selected surface area, 400 source points were randomly sampled and distributed uniformly across the surface. This allows X-rays to effectively cover the scanned object from multiple angles, thereby providing ample data support for reconstruction. Specifically, based on Equation~\eqref{eq:myequation20}, the angle $\theta$ is randomly selected from a uniform distribution within the range of $0$ to $2\pi$, while the angle $\phi$ is uniformly selected from a tilt angle range of $±10^\circ$. Finally, starting from the first point, we apply the nearest neighbor algorithm to reorder these points into a trajectory, as illustrated in Figure~\ref{RandomOrbit}.

\begin{figure}[!t]
    \centering
    \subfloat[]{\includegraphics[width=1.5in]{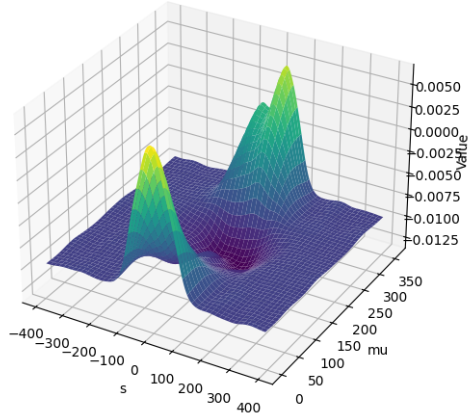}}
    \subfloat[]{\includegraphics[width=1.5in]{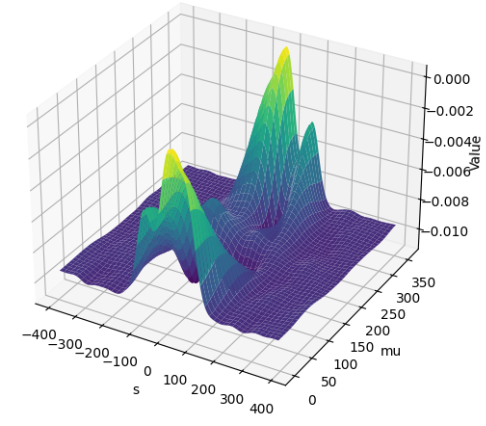}}
    \subfloat[]{\includegraphics[width=1.5in]{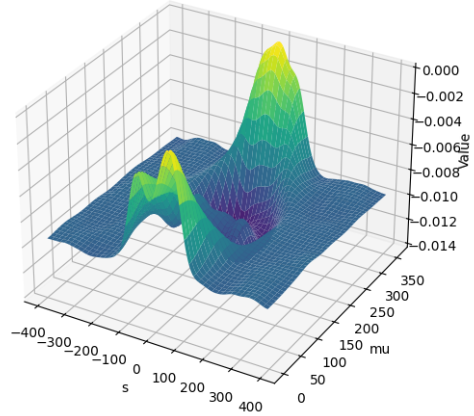}}
    \subfloat[]{\includegraphics[width=1.5in]{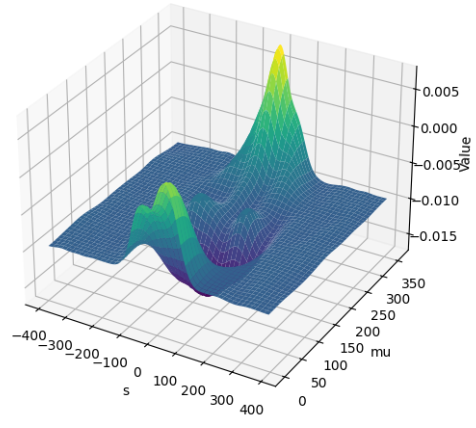}}
    \caption{Learned redundancy weights (Random Nearest Neighbor Orbit). (a)$\lambda = 0$. (b) $\lambda = 59$. (c)$\lambda = 119$. (d) $\lambda = 179$.}
    \label{learned_weight(random orbit)}
\end{figure}

A learning rate of $0.1$ was applied, and the network was trained for a total of 394 epochs, resulting in successful convergence. The learned redundancy weights obtained from this training are illustrated in Figure~\ref{learned_weight(random orbit)}.

\begin{figure}[!t]
\centering
\includegraphics[width=5.5in]{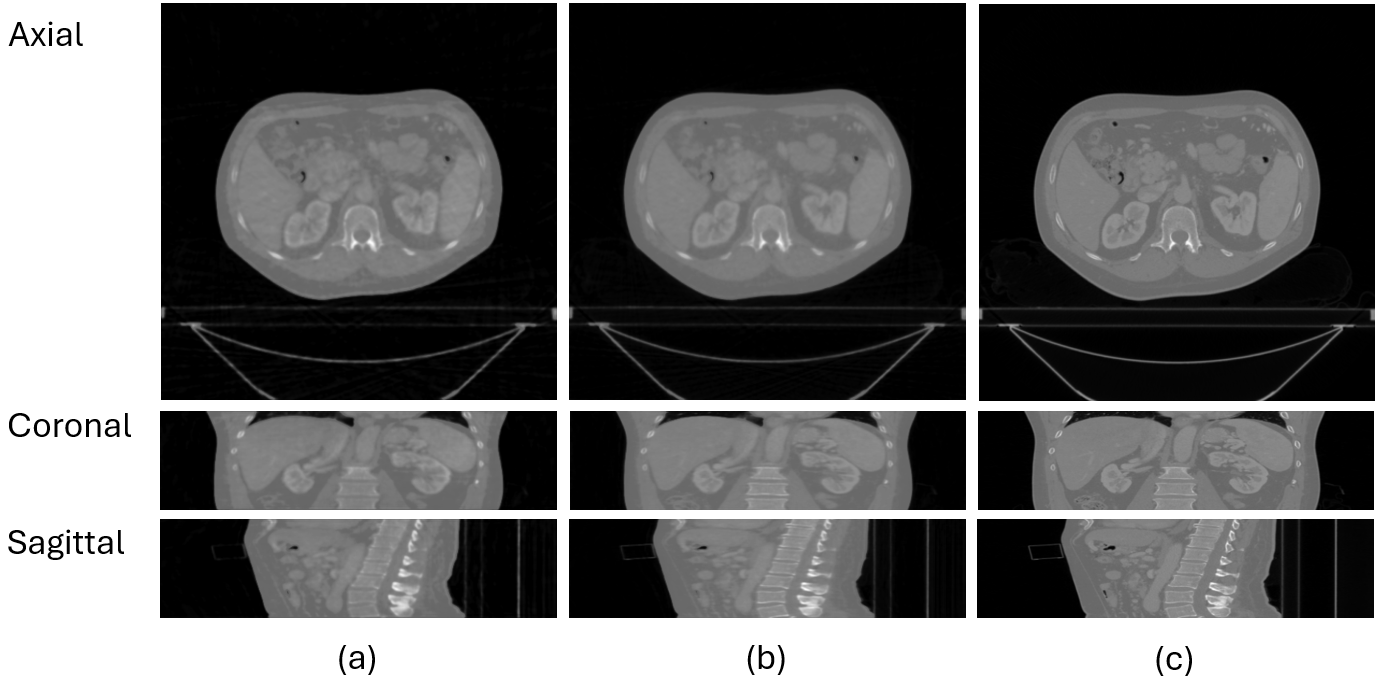}
\caption{Reconstructed results for the network(Random Nearest Neighbor Orbit). (a) Reconstruction using learned weights. (b) Iterative reconstruction result (300 iterations). (c) Ground truth.}
\label{result(random orbit)}
\end{figure}

\begin{table}[htbp]
\centering
\small
\caption{Comparison of Image Quality Metrics (Random Nearest Neighbor Orbit) with Mean ± Standard Deviation}
\begin{tabular}{@{}lcccccc@{}}
\toprule
 & \textbf{MSE}$\downarrow$ & \textbf{PSNR (dB)}$\uparrow$ & \textbf{SSIM}$\uparrow$& \textbf{Time (s)}$\downarrow$\\ 
\midrule
Our & 0.1035 ± 0.0175&  36.03± 1.23& 0.9232±0.0062 & 4.5\\
AIR(300) & 0.1348± 0.0514&  34.30± 2.07& 0.9150±0.0322 & 191\\
\bottomrule
\end{tabular}%
\label{tab: Random Nearest Neighbor Orbit}
\end{table}

The trained weights were subsequently employed for image reconstruction. To evaluate the efficacy of the reconstruction process, we employed a series of quantitative metrics, including MSE, PSNR, and SSIM, as detailed in Table~\ref{tab: Random Nearest Neighbor Orbit}. The analysis indicates that the reconstructed results achieved satisfactory performance across the specified metrics. Nevertheless, an examination of the central slices of the reconstructed volume in Figure~\ref{result(random orbit)} reveals the presence of artifacts in the axial view. 

The reason is that, given the complexity of the trajectory, the structural complexity of the required redundancy weights must also increase accordingly. However, in the case of discrete data, it is essential that the variation of the redundancy weights be as smooth as possible; otherwise, the introduction of artifacts may result. The neural network is only capable of balancing fidelity and smoothness within the constraints of the optimisation objective.

\section{Discussions}
\label{sec:5}


In the reconstruction of arbitrary orbits, the differentiable shift-variant FBP neural network markedly accelerates the reconstruction process in comparison to traditional iterative algorithms. However, the generation of new simulated datasets and the retraining of the neural network are required for each new orbit geometry, which represents an additional overhead when the orbit changes frequently. 

In comparison to analytical solutions, our method is capable of providing approximate solutions when the orbit is discontinuous, although some artifacts may be observed. Furthermore, our data-driven approach is able to automatically estimate the optimal redundancy weights for reconstruction based on the complexity of the orbit and the optimization objectives. This feature enables our method to potentially optimize based on other imaging effects, such as noise suppression, reduction of metal artifacts, or addressing limited view issues.
\section{Conclusion}
\label{sec:6}


This research propose a differentiable shift-variant FBP neural network designed for arbitrary CBCT orbits reconstruction using known operator learning.

The findings of this study indicate that the differentiable shift-variant FBP neural network is capable of learning the requisite parameters for reconstruction based on projection data with circular, sinusoidal, circle plus arc orbit geometry, and random nearest neighbor orbit, and it consistently demonstrates robust performance in these reconstruction tasks. The results demonstrate that deep learning technology can effectively resolve the estimation of redundancy weight in the shift-variant FBP algorithm, making a valuable contribution to the field of CBCT reconstruction for non-circular orbits.

The practical implications of this research lie in the potential development of a novel approach to robotic C-arm CT imaging technology. This approach could enable the rapid reconstruction of data with customized orbits. This improvement would represent a significant advancement in the field of medical imaging, particularly in robotic C-arm CT imaging technology.

Future research will concentrate on further reducing the number of neural network parameters and developing rapid reconstruction algorithms for multi-orbit reconstruction based on this approach. Furthermore, more sophisticated deep learning techniques will be investigated to address artifacts resulting from the discontinuities of the orbit.

\section{Acknowledgments}
The authors gratefully acknowledge the scientific support and HPC resources provided by the Erlangen National High Performance Computing Center (NHR@FAU) of the FAU Erlangen-Nürnberg. 
\section{References}
\bibliographystyle{unsrt}

\end{document}